\documentclass[letterpaper, 10 pt, conference]{ieeeconf}  

\IEEEoverridecommandlockouts                              

\usepackage{graphics} 
\usepackage{epsfig} 
\usepackage{mathptmx} 
\usepackage{times} 
\usepackage{amsmath} 
\usepackage{amssymb}  
\usepackage{microtype}
\usepackage{bm}
\usepackage{amsmath}
\usepackage{threeparttable}
\usepackage{multirow}

\title{\LARGE \bf
ROD: RGB-Only Fast and Efficient Off-road Freespace Detection
}

\author{Tong Sun$^{1,2}$, Hongliang Ye$^{3}$, Jilin Mei$^{1,\dag}$, Liang Chen$^{1}$, Fangzhou Zhao$^{1}$, Leiqiang Zong$^{4}$, Yu Hu$^{1,5,\dag}$ 
\thanks{\ddag This work was supported by National Natural Science Foundation of China under Grant No.U23B2034, No.62203424, and No.62176250; and in part by the Innovation Program of Institute of Computing Technology, Chinese Academy of Sciences under Grant No. 2024000112. }
\thanks{$^{1}$Research Center for Intelligent Computing Systems, Institute of Computing Technology, Chinese Academy of Sciences, Beijing, 100190, China.}%
\thanks{$^{2}$University of Chinese Academy of Sciences, Beijing, 100190, China.}
\thanks{$^{3}$Astronomical Computing Research Center, Zhejiang Lab}
\thanks{$^{4}$Beijing Special Vehicle Academy, Beijing, 100072, China. }%
\thanks{$^{5}$School of Computer Science and Technology, University of Chinese Academy of Sciences, Beijing, 100190, China.}%
\thanks{$^{\dag}$Correspondence: Jilin Mei, Yu Hu, \{meijilin, huyu\}@ict.ac.cn}%
}

\begin{document}

\maketitle
\thispagestyle{empty}
\pagestyle{empty}

\begin{abstract}

Off-road freespace detection is more challenging than on-road scenarios because of the blurred boundaries of traversable areas. Previous state-of-the-art (SOTA) methods employ multi-modal fusion of RGB images and LiDAR data. However, due to the significant increase in inference time when calculating surface normal maps from LiDAR data, multi-modal methods are not suitable for real-time applications, particularly in real-world scenarios where higher FPS is required compared to slow navigation. This paper presents a novel RGB-only approach for off-road freespace detection, named ROD, eliminating the reliance on LiDAR data and its computational demands. Specifically, we utilize a pre-trained Vision Transformer (ViT) to extract rich features from RGB images. Additionally, we design a lightweight yet efficient decoder, which together improve both precision and inference speed. ROD establishes a new SOTA on ORFD and RELLIS-3D datasets, as well as an inference speed of 50 FPS, significantly outperforming prior models. Our code will be available at https://github.com/STLIFE97/offroad\_roadseg.

\end{abstract}

\section{INTRODUCTION}

%

\begin{figure}[tbp]
\centerline{\includegraphics[width=\linewidth]{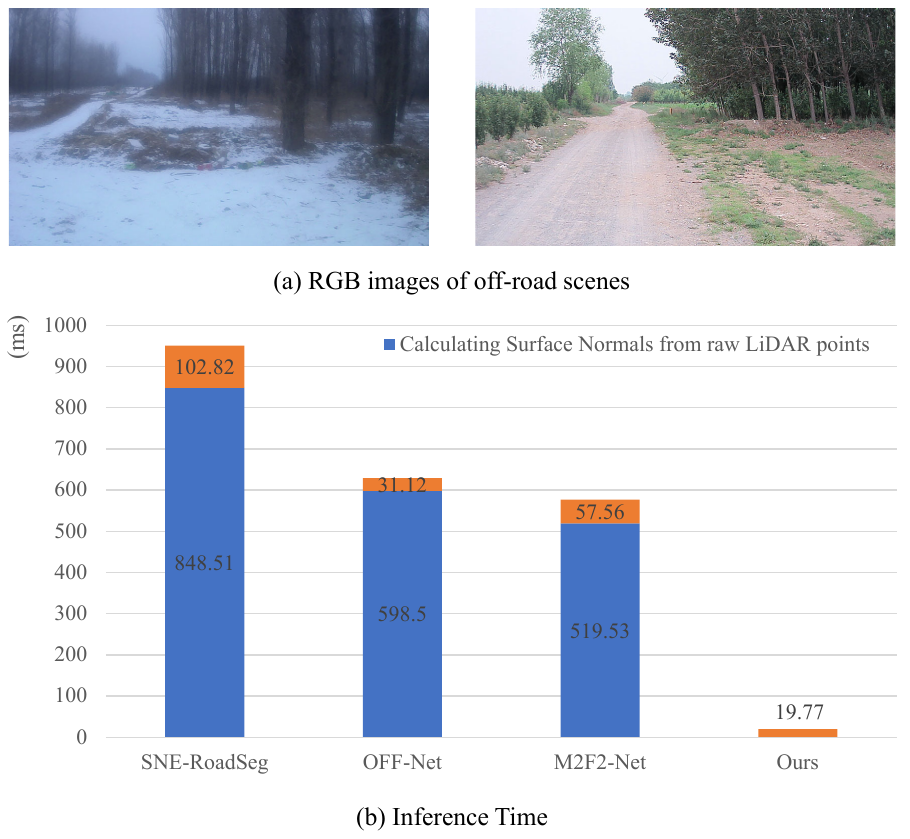}}
\caption{(a) provides examples of RGB images in off-road scenes, where the boundaries of traversable areas are not clearly defined. (b) illustrates the inference time for SNE-RoadSeg \cite{fan2020sne}, OFF-Net \cite{min2022orfd}, the prior SOTA method M2F2-Net\cite{ye2023m2f2} and ours. Multi-modal fusion based methods do not meet real-time requirements due to the computational steps involved in generating surface normal maps.}
\label{fig:cost}
\end{figure}

In recent years, autonomous driving is attracted growing attention and research. Freespace detection, a binary classification segmentation task, plays a fundamental role in autonomous driving systems by delineating navigable areas crucial for vehicle planning and control. The majority of research are mainly focused on urban on-road scenarios, characterized by well-defined features such as lanes and traffic signs \cite{fan2020sne}\cite{wang2021sne}. In contrast, research on off-road scenarios receives less attention \cite{wigness2019rugd}. These off-road scenarios present a higher degree of complexity and diversity, with the freespace being less distinct \cite{min2022orfd}. Vehicles are required to traverse a variety of terrains, including grasslands, sandy areas, icy grounds, snowy regions, and muddy terrains. As illustrated in Fig.~\ref{fig:cost}(a), the boundaries of traversable areas in off-road scenarios are blurred.

To cope with the above difficulties, previous SOTA methods predominantly rely on the multi-modal fusion of RGB imagery with LiDAR data \cite{ye2023m2f2}\cite{li2023roadformer}. While RGB images focus more on surface color, texture, and other visual information, LiDAR data focuses more on distance, depth, and position \cite{wang2022cross}. Multi-modal fusion can help these two modalities complement each other to achieve better performance, and thus multi-modal methods typically achieve higher accuracy \cite{huang2022multi}. Nonetheless, most methods for fusing LiDAR data require the calculation of surface normal maps \cite{ye2023m2f2}. which is a time-consuming process. This huge computational overhead significantly affects the inference speed, making the algorithm less viable for real-time applications on vehicles. For high-speed autonomous systems such as off-road vehicles or drones operating in dynamic environments, achieving high frame rates is critical to ensure timely decision-making and collision avoidance. As shown in Fig.~\ref{fig:cost}(b), the inference time for the M2F2-Net \cite{ye2023m2f2} is predominantly consumed by the generation of surface normal maps from LiDAR data, accounting for 90.02\% of the total inference time. Such latency severely limits the practical deployment of LiDAR-based methods in scenarios requiring rapid response. Beyond computational constraints, LiDAR sensors also impose significant hardware costs and energy consumption.

To reduce reliance on surface normal map, this study proposes a novel approach that utilizes a pre-trained ViT model to extract features only from RGB images, thereby significantly enhancing the inference speed. With the development of ViT \cite{dosovitskiy2020image}, the performance of large vision models \cite{kirillov2023segany}\cite{xiong2023efficientsam} improves as more data becomes available for pre-training. As shown in Fig.~\ref{fig:encoder-decoder}, the ViT encoder, which is employed for feature extraction, is kept frozen. Only the simple yet powerful decoder is trained. The RGB images are fed into the ViT encoder to generate an image embedding and to extract features from the latent layers of the transformer encoder blocks. These features are then used for subsequent fusion, and finally fed into the decoder for feature integration and prediction, culminating in the generation of the final prediction mask.

Our contribution can be summarized as follows:

\begin{itemize}
\item We conduct an investigation into the factors affecting inference speed in freespace detection models and propose a novel method, ROD. ROD only utilizes RGB data. This method surpasses prior multi-modal fusion methods in terms of both accuracy and inference speed.
\item ROD integrates a pre-trained ViT model into the off-road freespace detection task, and design a powerful decoder that effectively merges image embeddings with latent features.
\item 

The proposed ROD achieves performance on the ORFD and RELLIS-3D Dataset, with 98.3\% F1\_score and 96.7\% IoU on the ORFD dataset, and 97.1\% F1\_score and 97.8\% Accuracy on the RELLIS-3D dataset. Additionally, ROD achieves an inference speed of 50 FPS, comfortably meeting real-time requirements.

\end{itemize}

The remainder of this paper is organized as follows: Section 2 reviews related work, Section 3 details our model, Section 4 presents experiments, and Section 5 concludes the study.

\section{RELATED WORKS}

\subsection{Freespace Detection}

Freespace detection involves assigning a label to each pixel in traversable regions, which is categorized under semantic segmentation. The majority of research in this area focuses on on-road scenarios \cite{fan2020sne}\cite{wang2021sne}. With the development of deep learning, freespace detection has expanded to include both single-modal and multi-modal approaches.

Single-modal methods utilize either RGB images or LiDAR data. CNNs are widely used for RGB data, as studies \cite{fan2021learning}\cite{liu2018segmentation} demonstrate their effectiveness in freespace detection. \cite{fan2021learning} develops a driving scene generator to augment training data, while \cite{liu2018segmentation} introduces the RPP(Residual network with Pyramid Pooling) model, combining full convolutional networks with residual and pyramid pooling.
Within LiDAR-based methods, the most approaches employ occupancy grids and scene flow to delineate obstacles. MotionNet \cite{wu2020motionnet} adopts a voxel-based representation, and PointMotionNet \cite{wang2022pointmotionnet} opts for a point-based representation. Nevertheless, single-modal models that rely on either LiDAR or RGB images are limited by the data type they process. This limitation reduces the breadth of information available to the model, therefore, leading to lower accuracy compared to multi-modal systems \cite{huang2021makes}.

The multi-modal methods for freespace detection fuse RGB images with LiDAR data. PETRv2 \cite{liu2023petrv2} and CVT \cite{wang2022cross} utilize multi-modal camera fusion to create Bird's Eye View (BEV) images for road segmentation. BiFNet \cite{li2021bifnet} introduces a bi-directional fusion network for integrating point cloud images with BEVs. RoadSeg \cite{capellier2021fusion} proposes a method that leverages LiDAR data for road segmentation. SNE-RoadSeg \cite{fan2020sne} enhances segmentation performance by fusing surface normal maps with RGB image features. Its successor, SNE-RoadSeg+ \cite{wang2021sne}, further refines the accuracy of surface normal estimation and improves the network's performance and speed. Building on these advancements, M2F2-Net \cite{ye2023m2f2} and RoadFormer \cite{li2023roadformer} achieve the SOTA performance in freespace detection.

However, the computation of surface normal maps from LiDAR data in multi-modal methods is time-consuming, resulting in an inability to meet real-time requirements. Therefore, this paper aims to find a method that utilizes only RGB images to improve inference speed and preserve precision.

\begin{figure*}[htbp]
\centerline{\includegraphics[width=0.9\textwidth]{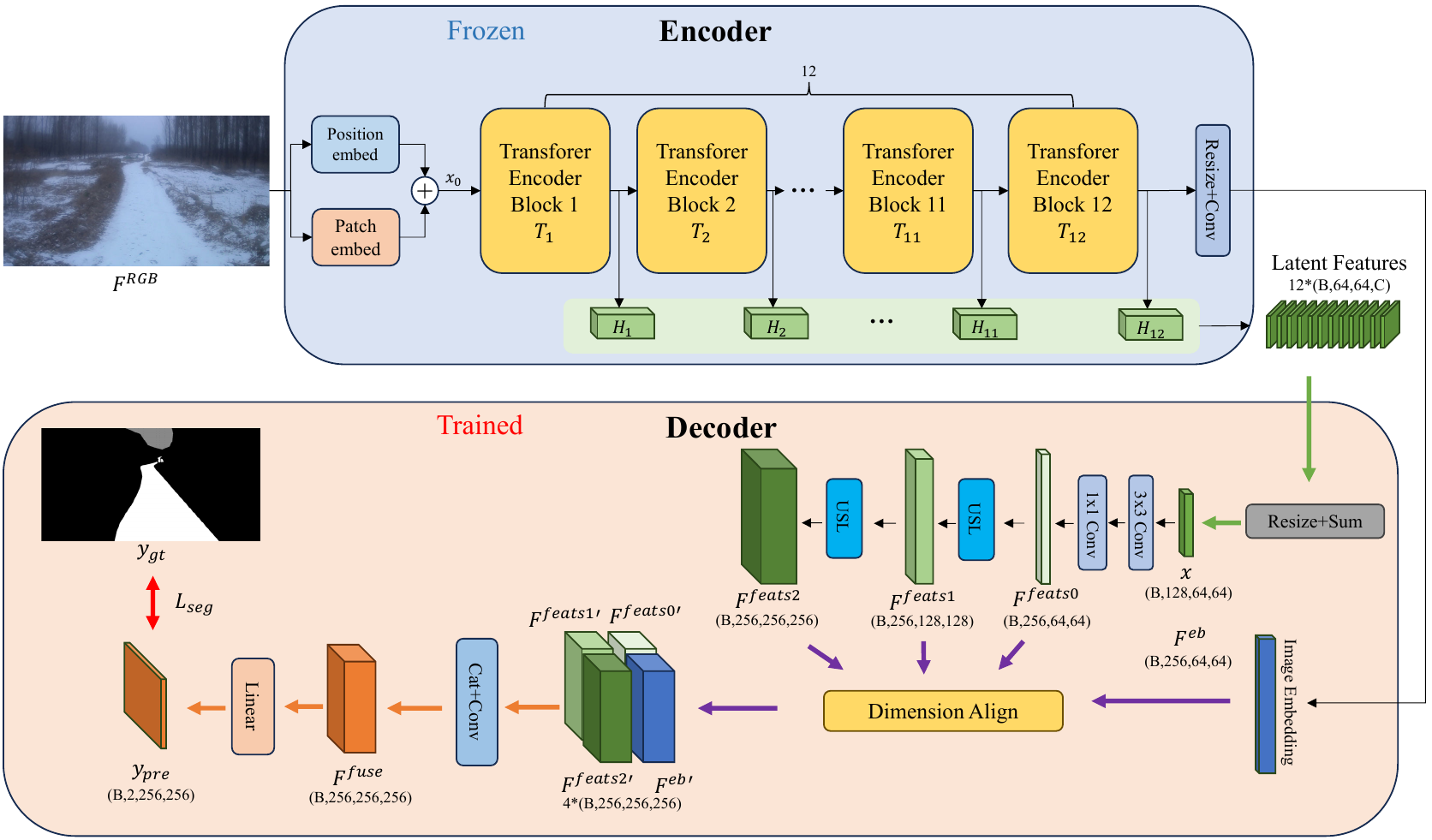}}
\caption{The detailed architecture of the network proposed in this paper. (a) the frozen encoder part takes RGB data as input and processes it through 12 layers of transformer encoder block $T_i$ to obtain the image embedding $F^{eb}$ and latent features $H_i$. The number of channels C in $H_i$ depends on the pre-trained model used, specifically, C=1280 for ViT-H, C=1024 for ViT-L, C=768 for ViT-B, C=384 for ViT-S and C=192 for ViT-T. (b) a simple but powerful decoder is designed to fuse the $H_i$ features and obtain features at different scales, then it outputs the prediction mask $y_{pre}$.}
\label{fig:encoder-decoder}
\end{figure*}

\subsection{Applications of Vision Transformer}

The Vision Transformer (ViT) \cite{dosovitskiy2020image} adapts the transformer, usually for NLP, to image classification. Its streamlined design allows scalability and, with sufficient pre-training data, can surpass CNNs in transferability across tasks \cite{dosovitskiy2020image}.This paves the way for the development of large-scale models. These models are successfully applied to an array of visual tasks, further substantiating ViT's prowess in feature extraction. For instance, ViT demonstrates exceptional performance in segmentation applications \cite{cen2023sad}\cite{chen2023sam_adapter}\cite{deng2023segment}, image inpainting \cite{yu2023inpaint}, image editing \cite{gao2023editanything}, object detection \cite{kirillov2023segany}, image captioning \cite{wang2023caption}, object tracking \cite{cheng2023segmenttrack}\cite{yang2023track}, and 3D object reconstruction \cite{shen2023anything}, among others.

Therefore, this paper attempts to integrate pre-train ViT model with freespace detection. However, a challenge with ViT is their large model size, which presents significant challenges for real-time applications, recent developments offer solutions. Lightweight ViT models such as MobileSAM \cite{zhang2023faster}, FastSAM \cite{zhao2023fast} are proposed, they increase the inference speed but decrease the accuracy. Subsequently EfficientSAM \cite{xiong2023efficientsam} introduces a method called SAM-leveraged masked image pretraining (SAMI), which results in a lightweight ViT encoder. With less than 5\% parameter of the original SAM, the performance drop is minimal, making it feasible for application in autonomous driving.

Thus, this study adopts the pre-trained ViT encoder from EfficientSAM for freespace detection. Leveraging the ViT encoder’s prowess in feature extraction, we are able to achieve high accuracy using only RGB images.

\subsection{Off-road Dataset}
Compared to a large number of on-road datasets, off-road datasets supporting freespace detection are few. The RUGD dataset \cite{wigness2019rugd} is designed for semantic understanding in off-road scenarios, including mountain trails, streams, parks, villages and so on. The RELLIS-3D \cite{jiang2021rellis} dataset is derived from RUGD and includes unique terrain like puddles. In addition, RELLIS-3D includes 3D LiDAR annotations. \cite{min2022orfd} proposed the ORFD dataset, which is an off-road freespace detection dataset. The dataset was collected under different scenarios (woodland, farmland, grassland, and countryside), different weather conditions (sunny, rainy, foggy and snowy), and different lighting conditions (bright light, daylight, dusk, and darkness). The proposed method will be tested on the RELLIS-3D \cite{jiang2021rellis} and ORFD \cite{min2022orfd} datasets.

\section{METHODOLOGY}

\subsection{Problem Definition} 
Freespace detection is a pixel-level classification problem, also known as semantic segmentation. The network $f$ classifies each pixel or point to determine if it is passable. Given one RGB image frame $F^{RGB}$ and its ground-truth mask $y_{gt}$ with pixel-level annotations, the goal is to optimize the model parameters of $f$ using the cross-entropy loss $L_{seg}$ :

\begin{equation}
    L_{seg} = - \sum_{H,W}^{} y_{gt}log(f(F^{RGB}))
\end{equation}
where $H$ and $W$ represent the dimensions of the RGB image.

\subsection{Overall Network Structure}

The network architecture, as detailed in Fig.~\ref{fig:encoder-decoder}, is composed of two parts: the frozen encoder, which utilizes a pre-trained ViT-S model \cite{xiong2023efficientsam} to extract features from RGB images; and the trained decoder, which fuses these features to produce the prediction mask and updates its parameters using cross-entropy loss.

The RGB image $F^{RGB}$ is utilized by the patch embed and position embed modules to generate positional and image patch encodings. These encodings are summed and passed through a 12-layer transformer encoder block $T_i$, obtaining 12 features from the latent layers $H_i$ and an image embedding $F^{eb}$. In the subsequent decode, the 12 feature maps $H_i$ are merged to obtain the $x$. The channel dimension of $x$ is expanded to create $F^{feats0}$, which is upsampled to produce $F^{feats1}$. $F^{feats1}$ is then upsampled again to generate $F^{feats2}$. The features $F^{feats0}$, $F^{feats1}$, $F^{feats2}$, and $F^{eb}$ are aligned and fused based on the dimensions of $F^{feats2}$ to obtain $F^{fuse}$. Lastly, $F^{fuse}$ is input to a linear layer to obtain the final prediction mask $y_{pre}$.

\subsection{Architecture of the Encoder}

The ViT encoder is utilized to extract high-performance features from only RGB images. A lightweight encoder, pre-trained by EfficientSAM \cite{xiong2023efficientsam}, is employed. The encoder's details are shown in Fig.~\ref{fig:encoder-decoder}. The encoder processes the input $F^{RGB}$, and subsequently, it generates image embeddings $F^{eb}$ and a set of latent features $H_i$.

Firstly, the RGB image $F^{RGB}$ is processed by the Patch Embed module $(patch)$, which segments it into a series of $16$x$16$ pixel patches as embedded representations; meanwhile, $F^{RGB}$ is fed to the Position Embed module $(pos)$ to obtain positional embeddings and ensure that the positional embeddings match the dimensions of the input image. The resulting outputs are combined to form the initial feature vector $x_0$, as detailed by the following equation:

\begin{equation}
    x_{0} = pos(F^{RGB}) + patch(F^{RGB})
\end{equation}

The ViT encoder sets the depth of transformer encoder blocks $T_i$ to 12 layers ($i = 12$). $x_0$ is fed into the first block $T_1$. Subsequently, each block $T_i$ receives the output from the preceding block as its input. with the formula as follows:

\begin{equation}
    H_{i} = \begin{cases}
     T_{i} (x_{0}) & \text{ if } i=1 \\
      T_{i} (H_{i-1}) & \text{ if } 1<i\le 12
    \end{cases}
\end{equation}

The ViT encoder processes data through 12 transformer blocks, resulting in 12 feature maps $H_i$. The last feature map $H_{12}$ input into $Resize+Conv$ module, undergoes a dimension permute, and is passed through two convolutional layers (1×1 and 3×3 kernels) to produce the final image embedding $F^{eb}$, sized (B, 256, 64, 64).

\subsection{Architecture of the Decoder}

\begin{figure}[tbp]
\centerline{\includegraphics[width=0.8\linewidth]{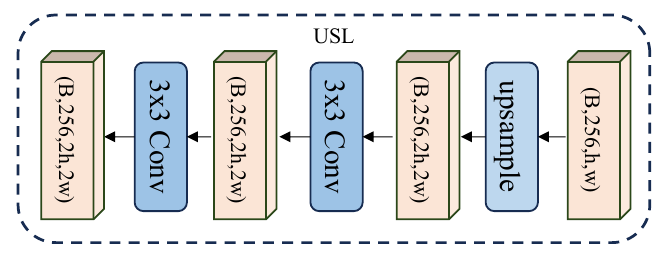}}
\caption{The specific structure of the upsample layers (USL). where $h$ and $w$ represent the dimensions of the feature map.}
\label{fig:upsample}
\end{figure}

To balance precision and inference speed, we design a simple yet powerful seg decoder to predict segmentation masks for freespace detection. The architecture of the decoder is detailed in Fig.~\ref{fig:encoder-decoder}.

Firstly, we permute the dimensions of the 12 latent features $H_i$ and downsampling them to (B, 128, 64, 64). Then summing these features to $\sum_{i=1}^{12} H_i$, and inputting the $\sum_{i=1}^{12} H_i$ through two consecutive convolutional layers, each with a 3×3 kernel, then the outputs is merged back with $\sum_{i=1}^{12} H_i$, creating a residual connection, and obtaining the summed features $x$. The formula is as follows:

\begin{equation}
    x  = \text{Conv}(\text{Conv}(\sum_{i=1}^{12} H_i)) + \sum_{i=1}^{12} H_i
\end{equation}

Fig.~\ref{fig:upsample} illustrates the specific structures of the upsample layers ($USL$).
The strategy consists of a sequence of two convolutional layers followed by a residual connection to retain finer details during the feature map's expansion. $x$ is passed through a 1×1 convolution, which doubles the channel count from 128 to 256, obtaining the feature $F^{feats0}$. The features $F^{feats0}$ are input into the upsample layers ($USL$) for bilinear upsample followed by two convolutional layers with a kernel size of 3×3, expanding the feature dimensions to 128×128 from 64×64. The output is merged with the upsampled $F^{feats0}$ to produce $F^{feats1}$. This process is replicated on $F^{feats1}$, scaling the features from 128×128 to 256×256, to yield the feature $F^{feats2}$. The formula are as follows:

\begin{equation}
    F^{feats0}  = \text{Conv}(x)
\end{equation}

\begin{equation}
    F^{feats1} = \text{USL}(F^{feats0}) + \text{upsample}(F^{feats0})
\end{equation}

\begin{equation}
    F^{feats2} = \text{USL}(F^{feats1}) + \text{upsample}(F^{feats1})
\end{equation}

Subsequently, the image embedding $F^{eb}$ and the feature maps $F^{feats0}$, $F^{feats1}$, and $F^{feats2}$ are upsampled to match dimensions of (B, 256, 256, 256) through bilinear interpolation, resulting in $F^{eb'}$, $F^{feats0'}$, $F^{feats1'}$, and $F^{feats2'}$. The formula is as follows:

\begin{align}
    &F^{eb'},F^{feats0'},F^{feats1'},F^{feats2'} \notag \\
    &= \text{interpolate}(F^{eb},F^{feats0},F^{feats1},F^{feats2})
\end{align}

Ultimately, the feature maps $F^{eb'}$, $F^{feats0'}$, $F^{feats1'}$, and $F^{feats2'}$ are concatenated to the dimensions (B, 4×256, 256, 256). This output is then processed by a 1×1 convolutional layer to downscale the channel number from 4×256 to 256, resulting $F^{fuse}$. Subsequently, a linear layer is applied to $F^{fuse}$ to produce the final predicted mask $y_{pre}$.

\begin{table*}[ht]
\renewcommand\arraystretch{1.5}
\caption{COMPARISON TO OTHER METHODS ON THE TESTING SET OF ORFD}
\label{tab:ORFD}
\begin{center}

\begin{tabular}{|c|c|c|c|c|c|c|c|cc|c|}
\hline
\multirow{2}{*}{Method}                             & \multirow{2}{*}{Modality} &\multirow{2}{*}{Year} & \multirow{2}{*}{Accuracy} & \multirow{2}{*}{Precision} & \multirow{2}{*}{Recall} & \multirow{2}{*}{F1\_score} & \multirow{2}{*}{IoU} & \multicolumn{2}{c|}{Inference Time (ms)}            & \multirow{2}{*}{FPS(total)} \\ \cline{9-10}
                                                    &                           &         &                  &                            &                         &                            &                      & \multicolumn{1}{c|}{CSN$^*$}    &
                                                 Model   Inference  &                             \\ \hline
U-Net \cite{ronneberger2015u}                                              & RGB     &  2015                & 0.959                     & 0.637                      & 0.537                   & 0.583                      & 0.411                & \multicolumn{1}{c|}{\textbf{0}} & 225.23            & 4.44                        \\ \hline
DLV3+-MNet \cite{chen2018encoder}                            & RGB    &2018                   & 0.976                     & 0.743                      & 0.847                   & 0.792                      & 0.655                & \multicolumn{1}{c|}{\textbf{0}} & \textbf{14.77}    & \textbf{67.70}              \\ \hline
DLV3+-R101 \cite{chen2018encoder}                             & RGB       &2018                & 0.980                     & 0.781                      & 0.871                   & 0.824                      & 0.700                & \multicolumn{1}{c|}{\textbf{0}} & 25.81             & 38.73                       \\ \hline
FuseNet \cite{hazirbas2017fusenet} & RGB+LiDAR       &2017          & 0.874                     & 0.745                      & 0.852                   & 0.795                      & 0.660                & \multicolumn{1}{c|}{-}          & -                 & -                           \\ \hline
MFNet \cite{ha2017mfnet}           & RGB+LiDAR     &2017            & -                         & 0.896                      & 0.903                   & 0.899                      & 0.817                & \multicolumn{1}{c|}{-}          & -                 & -                           \\ \hline
RTFNet \cite{sun2019rtfnet}        & RGB+LiDAR      &2019           & -                         & 0.842                      & 0.967                   & 0.900                      & 0.818                & \multicolumn{1}{c|}{-}          & -                 & -                           \\ \hline
SNE-RoadSeg \cite{fan2020sne}      & RGB+LiDAR      &2020           & 0.938                     & 0.867                      & 0.927                   & 0.896                      & 0.812                & \multicolumn{1}{c|}{848.51}     & 102.82            & 1.05                        \\ \hline
OFF-Net \cite{min2022orfd}         & RGB+LiDAR       &2022          & 0.945                     & 0.866                      & 0.943                   & 0.903                      & 0.823                & \multicolumn{1}{c|}{598.50}     & 31.12            & 1.58                        \\ \hline
M2F2-Net \cite{ye2023m2f2}         & RGB+LiDAR    &2023             & 0.981                     & 0.973                      & 0.955                   & 0.964                      & 0.931                & \multicolumn{1}{c|}{519.53}     & 57.56            & 1.73                        \\ \hline
RoadFormer \cite{li2023roadformer} & RGB+LiDAR      &2023           & 0.979                     & 0.951                      & 0.972                   & 0.961                      & 0.925                & \multicolumn{1}{c|}{-}          & -                 & -                           \\ \hline
Ours                                                & RGB       &2024                & \textbf{0.991}            & \textbf{0.984}             & \textbf{0.983}          & \textbf{0.983}             & \textbf{0.967}       & \multicolumn{1}{c|}{\textbf{0}} & 19.77             & 50.56                       \\ \hline
\multicolumn{10}{l}{$^{\mathrm{*}}$ Calculating Surface Normal(CSN) from raw LiDAR points.}\\
\multicolumn{10}{l}{${\mathrm{-}}$ The necessary code related to this is not provided in related paper.}\\
\end{tabular}
\end{center}
\end{table*}

\begin{table*}[ht]
\renewcommand\arraystretch{1.5}
\caption{COMPARISON TO OTHER METHODS ON THE TESTING SET OF RELLIS-3D}
\label{tab:RELLIS-3D}
\begin{center}
\begin{tabular}{|c|c|c|c|c|c|c|}
\hline
Method & Modality & Year & Accuracy & Precision & Recall & F1\_score  \\
\hline
U-Net \cite{ronneberger2015u} & RGB & 2015 & 0.977 & 0.809 & 0.856 & 0.832  \\
\hline
DLV3+-MNet \cite{chen2018encoder} & RGB & 2018 & 0.966 & 0.570 & 0.856 & 0.832  \\
\hline
DLV3+-R101 \cite{chen2018encoder} & RGB & 2018 & 0.966 & 0.586 & 0.428 & 0.495  \\
\hline
Real-NVP \cite{wellhausen2020safe} & RGB+LiDAR & 2020 & 0.5625 & 0.5710 & \begin{math} \mathbf{0.9742}\end{math} & 0.7001  \\
\hline
AE Based \cite{schmid2022self} & RGB+LiDAR & 2022 & 0.7348 & 0.7079 & 0.9181 & 0.7437  \\
\hline
Self-Supervisions Only \cite{seo2023learning} & RGB+LiDAR & 2023 & 0.9036 & 0.9164 & 0.8508 & 0.8622  \\
\hline
M2F2-Net \cite{ye2023m2f2} & RGB+LiDAR & 
2023 & 0.955 & 0.925 & 0.964 & 0.944  \\
\hline
Ours & RGB & 2024 & \begin{math} \mathbf{0.9786}\end{math} & \begin{math} \mathbf{0.9702}\end{math} & 0.9721 & \begin{math} \mathbf{0.9712}\end{math} \\
\hline

\end{tabular}
\end{center}
\end{table*}

\section{Experiment}

\begin{table}[!ht]
  \renewcommand\arraystretch{1.5}
    \caption{Ablation Study on ViT Encoder Selection}\label{tab:Pre-Trained Models}
    \centering
    \begin{threeparttable}          
      \begin{tabular}{|c|c|c|c|c|}
        \hline
        ViT Encoder             & Params (M)   & IoU & F1\_score      & FPS         \\ \hline
        ViT-H                     & 632      & 0.937    & 0.968          & 2.54           \\ \hline
        ViT-L                     & 308      & 0.933    & 0.965          & 4.58           \\ \hline
        ViT-B                     & 91      & 0.931    & 0.964          & 13.20           \\ \hline
        ViT-T           & \textbf{9.8}      & 0.927    & 0.962          & 46.34 \\ \hline
        ViT-S            & 25               & \textbf{0.967}     & \textbf{0.983} & \textbf{50.56} \\ \hline
        \end{tabular}
         \begin{tablenotes}    
        \footnotesize               
        \item[1] ViT-H is the pre-trained model sam\_vit\_h\_4b8939 in SAM \cite{kirillov2023segany};
        \item[2] ViT-L is the pre-trained model sam\_vit\_l\_0b3195 in SAM \cite{kirillov2023segany}; 
        \item[3] ViT-B is the pre-trained model sam\_vit\_b\_01ec64 in SAM \cite{kirillov2023segany}; 
        \item[4] ViT-T is the pre-trained model efficient\_sam\_vitt in EfficientSAM \cite{xiong2023efficientsam};       
        \item [5]  ViT-S is the pre-trained model efficient\_sam\_vits in EfficientSAM \cite{xiong2023efficientsam}. 
      \end{tablenotes}            
    \end{threeparttable}       
  \end{table}

\subsection{Experimental Settings}

\begin{figure*}[htbp]
\centerline{\includegraphics[width=0.7\textwidth]{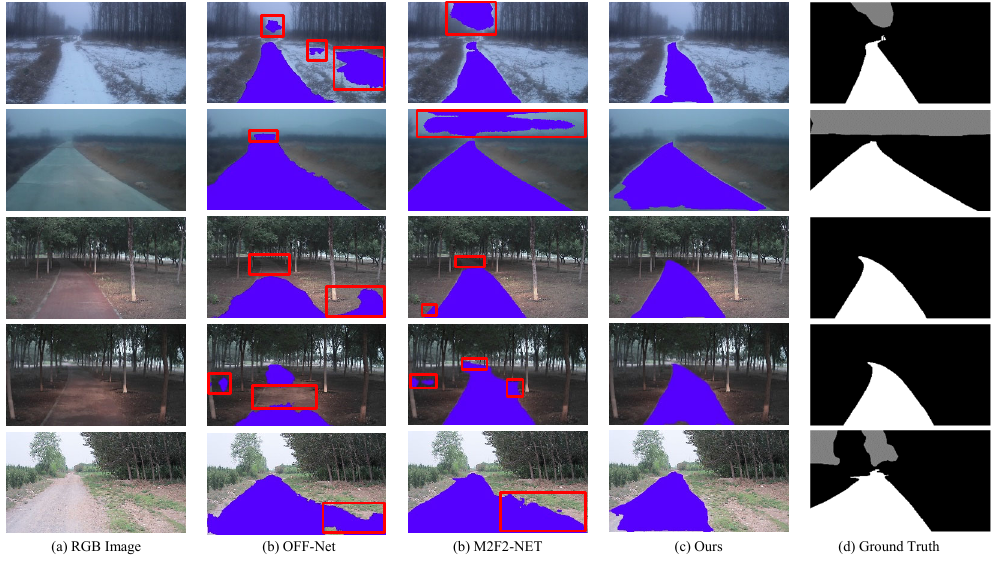}}

\caption{Qualitative results of OFF-Net \cite{min2022orfd}, M2F2-Net \cite{ye2023m2f2} and our method on ORFD dataset \cite{min2022orfd}. The red boxes are the area where OFF-Net and M2F2-Net predicts incorrectly but ours predicts correctly.}
\label{fig:vs}
\end{figure*}

Our method is evaluated on two datasets: the ORFD dataset \cite{min2022orfd}, which provides 12,198 pairs of RGB images and LiDAR point clouds; and the RELLIS-3D dataset \cite{jiang2021rellis}, a large-scale dataset focusing on freespace detection in off-road scenarios. The ORFD dataset contains 8,398 training samples, 1,245 validation samples, and 2,555 test samples, with image resolutions of 1280×720. In contrast, RELLIS-3D is a multi-modal dataset collected in off-road scenarios, comprising 13,556 LiDAR scans and 6,235 images. It includes 3,302 training samples, 983 validation samples, and 1,672 test samples, with images at a resolution of 1920×1200.

For testing on the ORFD dataset \cite{min2022orfd}, this paper uses the evaluation metrics from the M2F2-Net \cite{ye2023m2f2}, including IoU, Precision, Recall, Accuracy, F1\_score, and FPS.

For testing on the RELLIS-3D dataset \cite{jiang2021rellis}, this paper refers to the evaluation metrics from self-supervisions-only \cite{seo2023learning}, utilizing Precision (Pre), Recall (Rec), Accuracy (Acc), and F1\_score as the assessment criteria.

This study employs the AdamW optimizer with its default parameters, initiating the learning rate at 1e-3. We adopt `poly' decay policy for the learning rate, utilizing a decay power of 0.9. Batch size is set to 8. All experiments are conducted on a single NVIDIA RTX A6000 GPU.

\begin{figure*}[htbp]
\centerline{\includegraphics[width=0.8\textwidth]{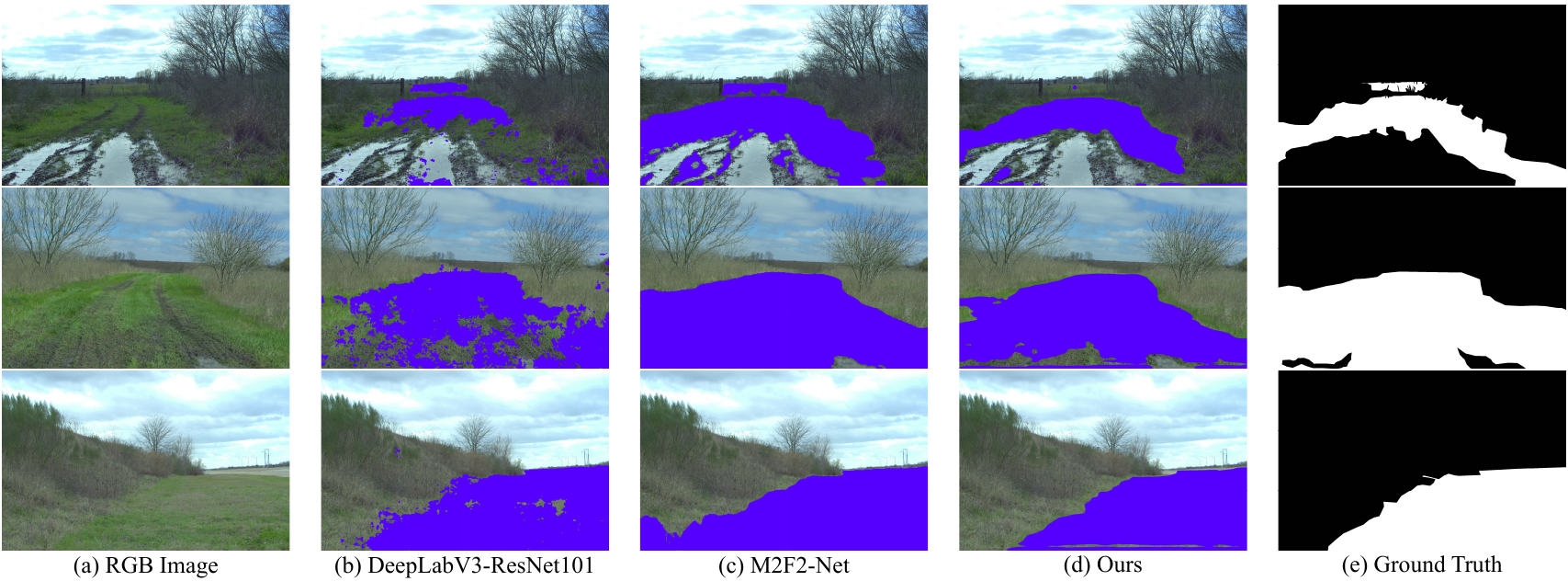}}

\caption{Qualitative results of DLV3+-R101 \cite{chen2018encoder}, M2F2-Net \cite{ye2023m2f2} and our method on RELLIS-3D dataset \cite{min2022orfd}. }
\label{fig:rellis}
\end{figure*}

\subsection{Comparisons on ORFD Dataset}

Comparisons on the ORFD dataset are presented in TABLE~\ref{tab:ORFD}. We compare our approach with various methods, including U-Net \cite{ronneberger2015u}, DeepLabV3Plus \cite{chen2018encoder} (DLV3+-MNet and DLV3+-R101), FuseNet \cite{hazirbas2017fusenet}, MFNet \cite{ha2017mfnet}, RTFNet \cite{sun2019rtfnet}, SNE-RoadSeg \cite{fan2020sne}, OFF-Net \cite{min2022orfd}, M2F2-Net \cite{ye2023m2f2}, and RoadFormer \cite{li2023roadformer}. Compared to methods that only use RGB \cite{ronneberger2015u}\cite{chen2018encoder}, surpasses DLV3+-R101 with a 15.9\% improvement in F1\_score and a 26.7\% improvement in IoU.  In methods that fuse RGB with LiDAR, M2F2-Net and RoadFormer are the previous SOTA methods for off-road freespace detection. Our method, which utilizes only the RGB images, surpasses M2F2-Net with a 0.6\% improvement in F1\_score and a 1.1\% improvement in IoU. In terms of inference speed, our method, by avoiding the calculation of surface normal maps, is 25 times faster than M2F2-Net.

As shown in Fig.~\ref{fig:vs}, we visualize the prediction results of our method and compare them with those of OFF-Net \cite{min2022orfd} and M2F2-Net \cite{ye2023m2f2} on the ORFD dataset. By comparing the predictions with Ground Truth, it becomes clear that our method outperforms OFF-Net and M2F2-Net. While all models excel in conditions with ample sunlight and clear weather. However, OFF-Net and M2F2-Net face several challenges in low light and adverse weather, including misidentifying the sky as ground, overlooking shaded passable roads, and incorrectly classifying roadside grass as part of the road.

\subsection{Comparisons on RELLIS-3D Dataset}

Comparisons on the RELLIS-3D dataset are detailed in Table~\ref{tab:RELLIS-3D}. We compare our approach with various methods, including U-Net \cite{ronneberger2015u}, DeepLabV3Plus \cite{chen2018encoder} (DLV3+-MNet and DLV3+-R101), Real-NVP \cite{wellhausen2020safe}, AE Based \cite{schmid2022self}, Self-Supervisions Only \cite{seo2023learning} and M2F2-Net \cite{ye2023m2f2}. The results indicate that our method outperforms these methods, achieving an 2.2\% improvement in Accuracy and a 2.7\% improvement in the F1\_score.

As shown in Fig.~\ref{fig:rellis}, we visualize the prediction results of our method and compare them with those of DeepLabV3Plus-ResNet101 \cite{chen2018encoder} (DLV3+-R101) and M2F2-Net \cite{ye2023m2f2} on the RELLIS-3D dataset. By comparing the predictions with Ground Truth, it becomes clear that our method outperforms DLV3+-R101 and M2F2-Net.

\subsection{Ablation Studies}

To determine the most suitable Vision Transformer (ViT) as the encoder for our feature extraction module, we evaluate ViT-H, ViT-L and ViT-B from SAM \cite{kirillov2023segany}, as well as ViT-T and ViT-S from EfficientSAM \cite{xiong2023efficientsam}, on the ORFD dataset \cite{min2022orfd}.  Our evaluations, as presented in Table~\ref{tab:Pre-Trained Models}, indicates that ViT-H's performance, with a speed of only 3 FPS, could not meet the criteria for real-time applications. According to \cite{xiong2023efficientsam}, instance segmentation models using ViT-T are 14\% faster than those based on ViT-S. However, this paper only utilizes the encoder parts of ViT-T and ViT-S, both of which have 12 blocks and output 12 hidden features for subsequent feature fusion, thus their inference speeds are roughly comparable. Consequently, we choose ViT-S as the encoder for subsequent experimental comparisons.

\section{CONCLUSIONS}

This paper introduces a fast and efficient method for off-road freespace detection. By employing a pre-trained ViT as the encoder, our approach effectively extracts features from RGB data alone, which enhances accuracy and significantly increases the model's inference speed by eliminating the need for surface normal map calculations. Additionally, we design a seg decoder that merges the encoder's latent features with image embeddings, ensuring greater detail retention and higher accuracy. Our method achieves SOTA performance on both the ORFD and RELLIS-3D datasets. Moreover, it performs inference at 50 FPS, significantly outperforming previous methods in terms of speed.






\bibliographystyle{IEEEtran}
\bibliography{IEEEabrv,IEEEexample}

\end{document}